\documentclass[5p, times]{elsarticle}

\makeatletter
\def\ps@pprintTitle{%
   \let\@oddhead\@empty
   \let\@evenhead\@empty
   \let\@oddfoot\@empty
   \let\@evenfoot\@oddfoot
}
\makeatother

\usepackage{bm}
\usepackage[svgnames,dvipsnames,table]{xcolor}
\usepackage[colorlinks]{hyperref}
\usepackage{amsmath}
\usepackage{amssymb}
\usepackage{adjustbox}
\usepackage{multirow}
\usepackage{tikz}
\usepackage{pgfplots}
\usepgfplotslibrary{patchplots}
\usetikzlibrary{trees,shapes,snakes,calc,matrix,positioning,fit,arrows,patterns,backgrounds,quotes,arrows.meta}
\usepackage{siunitx}
\usepackage{multicol}
\usepackage{subcaption} 
\usetikzlibrary{shapes.geometric}
\usetikzlibrary{shapes.arrows}
\usepgfplotslibrary{fillbetween}
\usepackage{array}
\usepackage{tabularx}
\usepackage{lipsum} 
 \usepackage{collcell}
\usepackage{hhline}
\usetikzlibrary{spy, calc}
\usepackage{array}
\usepackage{tabularx}
\usepackage{tabulary}
\usepackage{xspace}

\usepackage{bbding}
\usepackage{pifont}
\usepackage{diagbox}
\usepackage{booktabs}
\usepackage{makecell}
\usetikzlibrary{fit}
\usepackage{rotating}
\usepgfplotslibrary{colorbrewer}
\usepackage{fontawesome}
\usepackage{lscape}
\usepackage[para,online,flushleft]{threeparttable}
\usepackage{longtable}
\usepackage{tablefootnote}
\usepackage{makecell}
\usepackage{breakcites}
\usepackage{pifont}
\usepackage{hyphenat}
\begin{document}
\begin{frontmatter}

\title{PrunedCaps: A Case For Primary Capsules
Discrimination}


\author[1]{Ramin Sharifi}
\ead{raminsharifi@uvic.ca}
\author[1]{Pouya Shiri}
\ead{pouyashiri@uvic.ca}
\author[1]{Amirali Baniasadi}
\ead{amiralib@uvic.ca}


\address[1]{Department of Electrical and Computer Engineering, University of Victoria, 3800 Finnerty Rd, Victoria, BC, V8P 5J2 Canada}

\begin{abstract}
Capsule Networks (CapsNets) are a generation of
image classifiers with proven advantages over Convolutional
Neural Networks (CNNs). Better robustness to affine transfor-
mation and overlapping image detection are some of the benefits
associated with CapsNets. However, CapsNets cannot be classified
as a resource-efficient deep learning architecture due to the
high number of Primary Capsules (PCs). In addition, CapsNets’
training and testing are slow and resource hungry. This paper
investigates the possibility of Primary Capsules pruning in CapsNets on MNIST handwritten digits, Fashion-MNIST, CIFAR-10,
and SVHN datasets. We show that a pruned version of CapsNet
performs up to 9.90x faster than the conventional architecture by
removing 95\% percent of Capsules without a loss of accuracy.
Also, our pruned architecture saves on more than 95.36\% of
floating-point operations in the dynamic routing stage of the
architecture. Moreover, we provide insight into why some datasets
benefit significantly from pruning while others fall behind
\end{abstract}


\end{frontmatter}


\section{Introduction}
%

Artificial Neural Networks (ANNs) have recently become effective and popular models for the tasks of classification, pattern recognition, prediction and data clustering in various fields \cite{Abiodun2018}.     
ANNs are frequently used today for approximating functions due to properties such as adaptability, self-learning, non-linearity and advanced mapping of the inputs to the outputs \cite{Wang2018}. One of the important applications of ANNs is classification where data should be categorized into several classes. 

\par
Computational costs and memory intensity of ANNs makes their deployment in environments with limited resources difficult \cite{MLSYS2020_d2ddea18}.   
Complex networks usually achieve high accuracy at the cost of slow training and inference phases \cite{Carbin2019}. Neural Network approximation methods redesign complex ANNs with the goal of saving memory and reducing computation power. These improvements usually come at the cost of marginal accuracy losses. Such approximations play a crucial role in reducing the latency and improving throughput in various applications. These methods are categorised into two major groups: quantization and weight reduction. Quantization algorithms focus on reducing the precision of the network parameters. Weight reduction solutions eradicate redundant neurons or structures \cite{Wang2019}. 

\par
Pruning is one of the widely researched topics and an example of weight reduction methods. Pruning an ANN is the process of removing neurons from a network. This method aims to omit those neurons that have zero or insignificant impact on the output’s accuracy. ANN pruning was introduced by Lecun et al \cite{LeCun1990}. 

There are two main types of pruning for ANNs: fine-grain and coarse-grain. Similarity evaluations and iterative pruning processes both implement pruning at an element-wise granularity. These pruning techniques are referred to as fine-grain pruning. Fine-grain pruning can result in uneven weight distributions which can affect hardware’s throughput. Coarse-grained pruning techniques, however, can find the sweet spot in the compression-throughput trade-off by taking into account the whole architecture instead of focusing on individual elements\cite{Lebedev2016}.

\par
Sabour et al introduced Capsule Network (CapsNet) as the next generation of ANNs \cite{Sabour2017}. The computational unit of CapsNet consists of a vector of neurons referred to as a capsule. The final representation of this network is also in the form of capsules. CapsNet contains more information compared to other conventional ANNs such as Convolutional Neural Networks (CNNs). This is in part due to the absence of pooling layers. Pooling is the act of reducing the size of feature maps, which are used in conventional CNNs. Pooling layers in CNNs are the main cause of losing information. Hence, CapsNets, which do not use pooling layers preserves information more effectively.    
\par

\par

Primary Capsules (PC) are reshaped vectors calculated through multiplication of feature extractor's output and a weight matrix. PCs contain the most trainable number of parameters in the entire CapsNet architecture. Another trainable part of CapsNets are the convolutional layers which build the feature extractor and translate images to the feature maps. The last section which has trainable parameters is the decoder layer which reconstructs the image and compares it to the input. The focus of this paper is on pruning PCs to speed-up inference.

\par
Pooling layers are mainly used to reduce the size of feature maps by removing elements that do not play a significant role in deciding the output. Not using pooling layers may cause over-fitting in smaller Networks. Choosing the maximum element or averaging certain elements are just examples of pooling. Since CapNets do not utilize pooling layers, the accumulation of convolutional layer output results in a very large multi-dimensional matrix. Having a multi-dimensional matrix requires time consuming computations which affects the architecture's reliability for low latency applications such as self-driving cars. In this work we use a pruning technique that relies on first-order gradient information for parameter salience. This technique has been successful in CNNs, and as we show, delivers promising results in CapsNets, removing more than {95}\% of the parameters with a minuscule drop in accuracy.

\par 
In summary our contributions are as follows:
\begin{itemize}
    \item Pruning CapsNet’s Primary Capsules using an approach employing Taylor's expansion approximation. Primary Capsules are a reshaped representations of features which are multiplied by a matrix. We use Taylor's expansion as a metric to select and remove PCs which are redundant to the architecture. 
    \item A comprehensive analysis of experiments on popular datasets such as MNIST hand written digits, Fashion-MNIST, SVHN, CIFAR-10 and SmallNORB. We provide detailed analysis on variations in inference time, FLOPS counts for networks employing different number of capsules.
    \item Providing insight on how pruned CapsNets behave for different number of PCs. We also study how the dataset's complexity (i.e., size and features) impacts Capsnet's behaviour. We show how Taylor's expansion pruning removes PCs. In addition we report how pruning-enabled changes impacts CapsNets.
\end{itemize}

\par

The rest of this chapter is organized as follows. Section II describes related works. Section III details the background and our method. Section IV reports the results and discussions. Section V offers concluding remarks.

\section{Related Works}
There are several works focusing on pruning ANNs.
Lecun et al. and Hassibi et al. are the two pioneers in removing unnecessary weights in ANNs \cite{LeCun1990,Hassibi1993}.
Pruning can be effective in different ways. Han et al. and Suzuki et al. show how pruning can sometimes increase accuracy by longer training compared to the baseline model and without overfitting \cite{Han2015,Suzuki2018}.  Residual Network (ResNet) are a kind of Deep Learning Network which use residual blocks. Residual blocks help with having more layers to train without the problem of vanishing gradients by appending the input feature to the output of the block. Previous research has established that the error rate on the ResNet20 network can be reduced by setting sparse weights to the pruned network \cite{Leee2019}. Kalchbrenner et al. \cite{Kalchbrenner2018} use pruning for efficient audio synthesis. They use a single Recurrent Neural Network (RNN) referred to as WaveNet. Their findings suggest that a sparse architecture can outperform a smaller dense network with the same number of parameters. 
\par
Random pruning is the process of selecting the target neurons and removing them randomly. Researchers have observed that method-based pruning outperforms random pruning.
 \cite{Carbin2019,Gale2019,Yu2018,Pavlov2019}. Frankle et al. \cite{Carbin2019} argue that a large, dense, and randomly initialized networks contain subnetworks. These subnetworks can be trained to perform competitively compared to their parent network. These subnetworks are initialized with the original weights of the network.
\par
Lottery ticket pruning is the process of finding subnetworks with sizes  under 10-20\% of the size of the original network. Lee et al. \cite{Lee2019} have identified irrelevant connections using a method referred to as “SNIP” (Single-Shot Network Pruning). They start pruning prior to training. This can lead to better results due to network sparsity at initialization.  
\par
Pruning can also be done by removing neurons from all layers of a network in a uniform fashion. To date, several studies have investigated this approach. Performing pruning uniformly is outperformed by a smart parameter allocation technique \cite{Han2015,Gale2019}. Lou et al. \cite{Luo2017} demonstrate a compressed and accelerated pruning method for CNNs. Their method does not follow a uniform fashion and achieves better accuracy. 
\par
A pruned network will lose accuracy if it is not fine-tuned. Fine-tuning is the process of continuing the training of an ANN after initial prunning. Recent studies \cite{Zhang2016,Yu2018} suggest that if all weights are set to zero, training a pruned architecture falls behind fine-tuning.

\section{Background}

In this section, we review the background. First, we review CapsNet and its architecture. Afterwards, we explain the pruning techniques.
\subsection{Capsule Network (CapsNet)}
\begin{figure*}[h!]
\centerline{\includegraphics[scale=0.4]{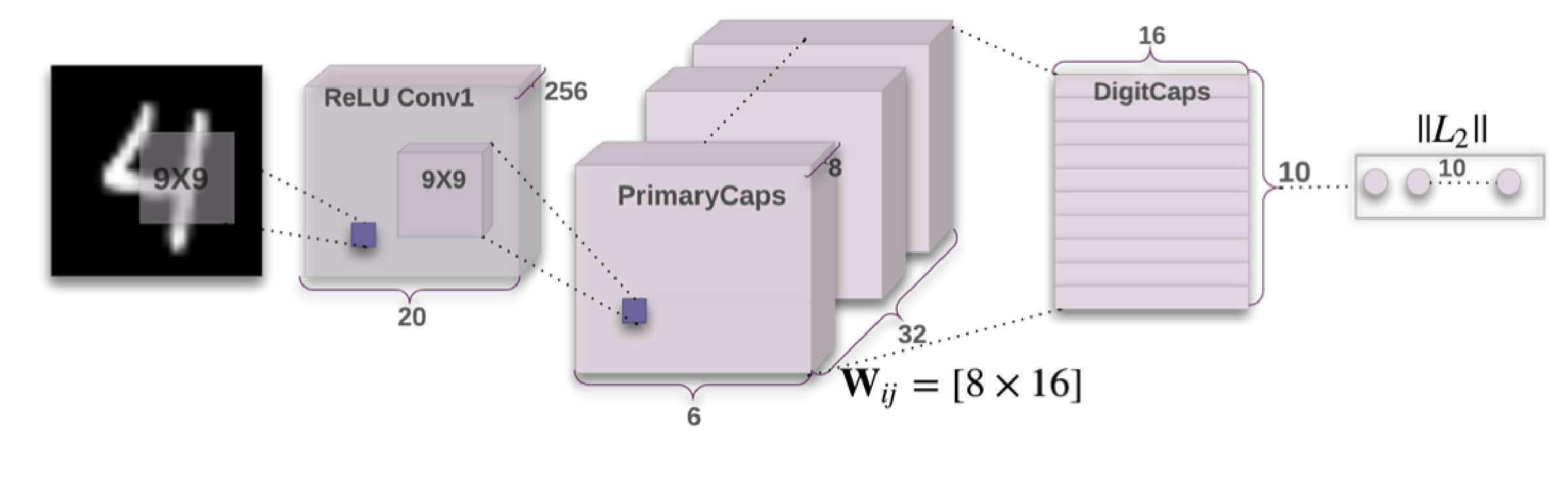}}
\caption{Original CapsNet Architecture.\cite{Sabour2017}}
\label{fig:Capsule Architecture}
\end{figure*}

The basic computational unit in CapsNet are capsules (vectors of neurons). The architecture of CapsNet is shown in figure \ref{fig:Capsule Architecture}. 
According to the figure, the network starts with extraction of low-level features using two convolutional layers. The extracted features are then reshaped to vectors. These vectors are then multiplied by a matrix, encoding the spatial relationship between them. The resulting vectors are referred to as Primary Capsules (PCs). 
\par
The next layer of capsules (the output capsules) are inferred from PCs. There is a fully-connected relationship between the PCs and the output capsules. However, the coefficients are not learned during the training process. Instead, they are determined in every iteration of the training process by an iterative algorithm referred to as Dynamic Routing (DR). This algorithm determines the contribution of each PC in each output capsule based on the level of agreement among PCs. 

\par
CapsNet is designed such that there are as many output capsules as the number of categories in the classification task. The output capsules hold two important pieces of information. First, the capsule with the highest magnitude is the capsule corresponding to the correct class. Second, the different dimensions of each output capsule holds information about instantiation parameters associated with the input image such as pose and deformation \cite{Sabour2017}.

CapsNet also includes a basic decoder which is used to reconstruct the input images using the output capsules as input. The decoder consists of fully-connected layers. The reconstructed image is compared to the input images. The Euclidean norm of difference between the two images is known as reconstruction loss, and used to regularize the training process.

The loss function in CapsNet consists of two terms. The first term is the reconstruction loss explained earlier. The second term is referred to as the margin loss. This term is based on the predictions made by the output capsules. There is a penalty considered for each output capsule based on the following equation:
\[ L_k = T_k \max(0,m^+-||V_k||)^2 + \lambda(1-T_k)\max(0,||V_k||-m^-)^2 \]
where $L_k$ denotes the loss term for each capsule, $T_k$ is determined based on the prediction of each capsule: it is set to one for correct predictions and to zero otherwise, $\lambda$ is the weight considered for penalizing wrong predictions, and $m^+$ and $m^-$ are used to remove capsules with high or low probabilities from taking part in the margin loss.

\subsection{Pruning Methods}
The process of pruning consists of removing the least important parameters. It is essential to minimize the difference in accuracy between the normal and the pruned networks. Optimally, the impact of removing each parameter should be evaluated by investigating how it could impact network accuracy.
\par
There are two criteria used in estimating the importance of the parameters. The first criterion is based on the signed change in the loss $L_{D,W'} - L_{D,W}$, where $L_{D,W}$ is the value of loss for the network over dataset $D$ using parameters $W$. $W'$ is the new set of parameters with the redundant parameters removed. The second criterion focuses on the absolute difference in the loss $|L_{D,W'} - L_{D,W}|$. 
\par
It is significantly expensive to consider the effect of each parameter individually on the dataset. Therefore there are various estimation methods including Minimum Weight, Activation, and Taylor Expansion. Here we briefly explain each method.

\subsubsection{Minimum Weight}
The Minimum Weight method is based on the magnitude of the parameters in the kernel. The intuition behind this method is the following: the lower $l2$-norm of a kernel is, the less important are the features it detects. Regularizing the network would be beneficial to this method since kernels corresponding to less important features are pushed to have smaller values.

\subsubsection{Activation}
The Rectified Linear Unit (ReLU) is a mathematical function defined as $max(0,x)$ where $x$ is the input of the function.
The ReLU activation function is sparse and is used as a feature detector in the convolutional layers.  This is due to the fact that convolutional layers check for existing features in their input. Any feature element with zero or negative activation by the Convolutional layer means that the associated feature does not exist.

Activation pruning works by removing small activation values. The issue with activation pruning is that it can only use the ReLU activation function.  Other common activation functions used in CNNs would not clip the negative values like ReLU, therefore we cannot use them for pruning the feature extractor.

\subsubsection{Taylor Expansion}
This pruning method aims to minimize $|L_{D,W'} - L_{D,W}|$ by removing some parameters\cite{Mal2017}. The Taylor expansion method can approximate the effect of removing a parameter on the loss function. Assuming that parameters are independent, for parameter $i$ our goal is  to minimize $|L_{D,W_i=0} - L_{D,W_i}|$. We can estimate the minimization target using the Taylor expansion. The following is the equation of Taylor expansion for function $f$ at the point $x=a$:
\begin{equation}
    f(x) = \sum_{0}^{\infty} \frac{f^{(n)}(a)}{n!}(x-a)^n
\end{equation} 
where $f^(n)$ denotes the n-th order derivative of function $f(x)$.
\par
Now, we use the Taylor expansion for function $L_{D,W_i=0}$ at $W_i=0$:
\begin{equation}
    L_{D,W_i=0} = L_{D,W_i} - \frac{\partial{L}}{\partial{W_i}}W_i + R
    \label{eq-taylor}
\end{equation}
where $R$ contains all the remaining higher order terms in the Taylor expansion. We neglect the remaining terms for two reasons. First, it increases the computation complexity. Second, the value of the higher order terms in the Taylor expansion are often negligible compared to the first order term. 
\par
Based on the equation \ref{eq-taylor}, the minimization target now changes:
\begin{equation}
    |L_{D,W_i=0} - L_{D,W_i}| = \left|\frac{\partial{L}}{\partial{W_i}}W_i\right|
\end{equation}

In short, the Taylor method results in pruning parameters associated with small gradients in the loss function. Implementing this method is feasible, as it needs the multiplication of the gradient of a parameter by the parameter itself. Such information are all available during back-propagation. Considering the effect of all parameters, the pruning method using Taylor expansion could be reformulated as a problem of minimizing the following function:
\begin{equation}
    F(W)= \left|\frac{1}{M}\sum_{m}\frac{\partial{L}}{\partial{W_m}}W_m \right|
\end{equation}
where M is the number of all parameters. 
\subsection{PrunedCaps Method}
\par
In this section, we use the Taylor's expansion method.
A Primary Capsule is a reshaped representation of the feature extraction layer which is multiplied by a matrix that encodes the spatial relations into vectors. Since PCs are multi-dimensional, changing them requires high computational power. 

Feature vectors which are ignored and not processed further into the network, can be removed from the network. This is despite the fact that they are computed every time there is an input to the network.
\par
Selecting which Primary Capsules to remove in each pruning epoch, requires ranking capsules according to their activation and back-propagation gradients.
Therefore, Primary Capsules are ranked according to the product of their back-propagation gradient and activation. 
The results of the activation times the gradients are then divided by the number of features present. If activation times the gradients results in a high number compared to the output of the same process for other PCs, then we can assume that it has high significance to the the network for inference. In our case, the number of features which are present is the number of remaining Primary Capsules.
This is the implementation that is presented in eq. (4). $F(W)$ is a weighted average of activations. The weights are gradients which represents the significance of the activation. 
\par
Our method prunes the Primary Capsules that show little to no change with respect to their output. As iterations grow toward a complete epoch, this value is accumulated giving us the result indicating how each Primary Capsule would behave over a dataset.
After each pruning epoch ends, rankings are sorted and the lowest values for the ranking criteria are selected. The number of PCs to be removed is a hyper-parameter which is set before the training begins.

\begin{figure}[h!]
\centerline{\includegraphics[scale=0.3]{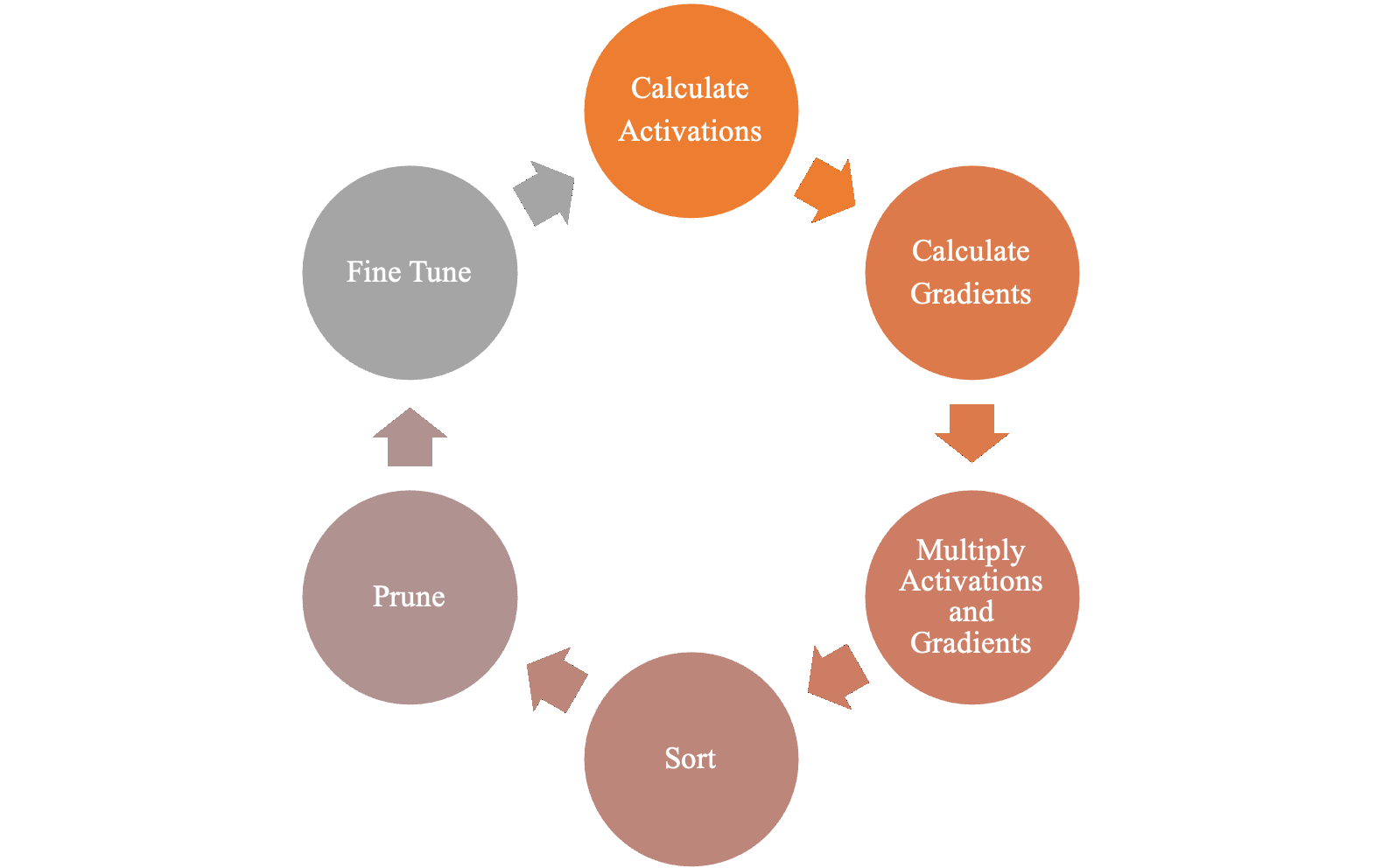}}
\caption{Summary of Pruning Algorithm}
\label{fig:algo}
\end{figure}

After each pruning phase, CapsNet goes through a training phase to restore its accuracy and adapt to the changes made to its architecture. This training phase is usually referred to as fine-tuning phase. 
Between pruning epochs, we fine-tune for several epochs for the network to reach its maximum possible accuracy. The number of fine-tuning epochs is set empirically. If we increase the number of epochs, there would be minuscule accuracy gain which can be neglected. If this number is reduced we cannot be sure if the network has reached its maximum accuracy. In our case, we have tested different number of epochs and decided on a number ( i.e., 50) which ensures a fully-trained network.  
We have summarized the algorithm in fig. \ref{fig:algo}. 

\section{Experiments and Results}
In this section, we present experiments and results. Experiments are done with a machine equipped with an NVIDIA 2080 Ti GPU and 32 GB of Memory and a machine equipped with an NVIDIA V100 GPU and 13 GB of Memory. 
\par
We use MNIST handwritten digits, Fashion-MNIST, CIFAR-10, SVHN and SmallNORB datasets. MNIST handwritten digits and Fashion MNIST datasets contain 28x28 single-channel images. MNIST handwritten digits contain images of handwritten digits from zero to nine. Fashion MNIST dataset includes images of different pieces of clothing. They both share the same number of classes for classification. 
 By default, MNIST handwritten digits and Fashion MNIST are divided into 50,000 and 10,000 images for training and testing. 
 CIFAR-10 and SVHN datasets share the same 32x32 image size. They both have ten classes of RGB images. CIFAR-10 has 50,000 images for the training set and 10,000 images for the testing set, whereas SVHN has 73,257 images for the training set and 26,032 images for the testing set. 
CIFAR-10 consists of ten very different classes: airplanes, automobiles, birds, cats, deer, dogs, frogs, horses, ships and trucks. SVHN dataset classes are the same as MNIST handwritten digits, but the digits are house numbers obtained from Google’s Street View \cite{37648}.
\par
SmallNORB images are much larger at 96x96 pixel compared to the previous datasets. SmallNORB uses single channel grayscale for color representation. SmallNORB consists of 48600 different images, in five different categories of toys which are: human, airplanes, trucks, cars and four-legged animals. Each toy has been photographed with different conditions. These conditions include: lightning, elevation and azimuth. \cite{smallnorb}

\begin{figure}[h!]
\centerline{\includegraphics[scale=0.5]{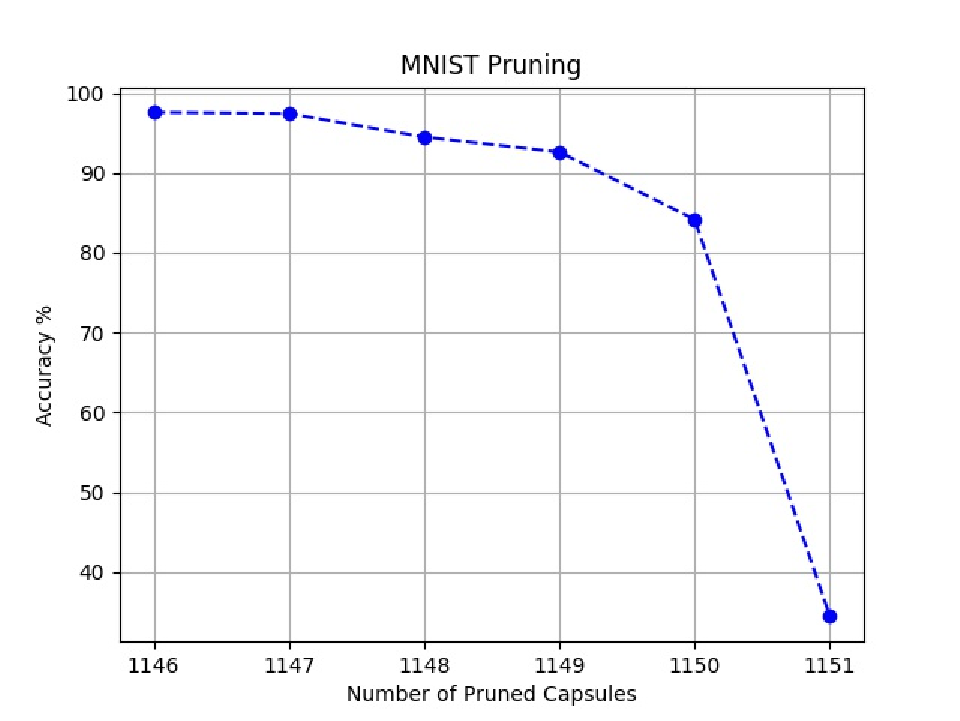}}
\caption{CapsNets accuracy drop on MNIST starts when only 5 PCs are remaining}
\label{fig:mnist}
\end{figure}
 For 28x28 sized images, we start with 1152 Primary Capsules, which is the baseline number of capsules in the original architecture. We start by training the network to reach its baseline accuracy on each dataset. The baseline architecture reaches 99.47 \% accuracy on MNIST and 90.23\% accuracy on Fashion MNIST test samples of the datasets. We save the weights at the end of the training phase so we can use them in the pruning phase. 
 
 \begin{figure}[h!]
\centerline{\includegraphics[scale=0.6]{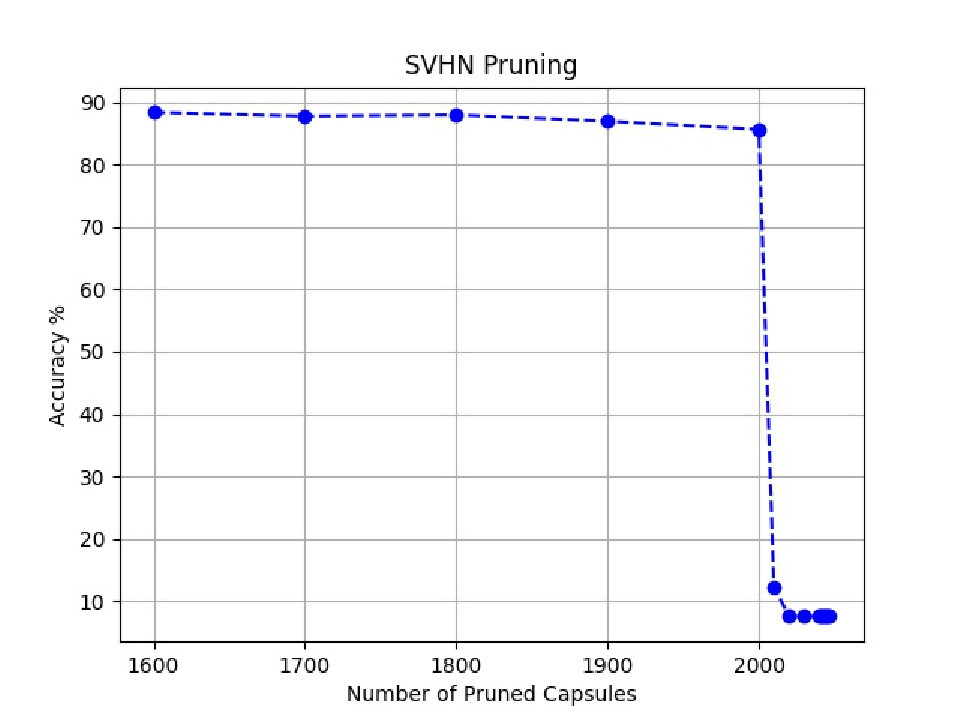}}
\caption{Minuscule drop of accuracy at 2000 PCs pruned on SVHN dataset}
\label{fig:svhn}
\end{figure}

 \par
\begin{figure}[h!]
\centerline{\includegraphics[scale=0.6]{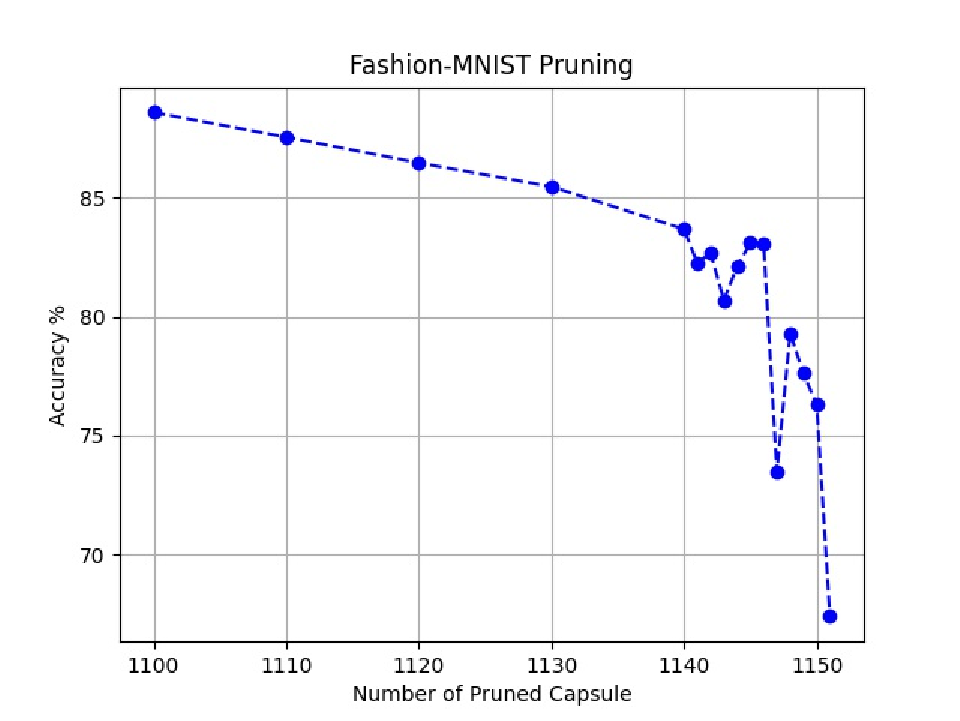}}
\caption{Accuracy drop starts when 52 PCs are remaining for Fashion-MNIST dataset. F-MNIST is considered to be more complex than MNIST.}
\label{fig:fmnist}
\end{figure}

Pruning weights start by one epoch of training so that the backpropagation gradients can be calculated for PC rankings. Since more than half of the PCs are either zero or have values near zero, we chose to prune 100 PCs during each epoch for the first 1100 PCs. After each pruning epoch, the network is fine-tuned on the dataset to reach its maximum accuracy. After 1100 PCs are pruned, we then lower the number of PCs to be pruned to 10 PCs. This change in the number of the pruned PCs is because the accuracy starts to decline at 1100 PCs pruned. As we show in Fig. \ref{fig:mnist} and \ref{fig:fmnist}, the network cannot recover from the PCs removed from its architecture for MNIST and FMNIST datasets beyond a certain point. This figure shows the maximum accuracy reached by the architecture at each accessible number of capsules.
\par
The same process is repeated for CIFAR-10 and SVHN datasets. The main difference is the starting number of Primary Capsules. The starting number of PCs is a function of the input's size. For MNIST and Fashion-MNIST this number is 1152 which is equal to 6x6x32. This is the reshaped feature size of the input image which has gone  through two layers of Convolutional layers: first layer of stride one and second layer of stride two. For 32x32 pixel images, 1152 changes to 2048 which is equal to 8x8x32. Sabour et al. discussed how input image shapes change throughout the CapsNet extensively \cite{Sabour2017}.
\par

\begin{figure}[h!]
\centerline{\includegraphics[width=0.5\textwidth, scale=0.6]{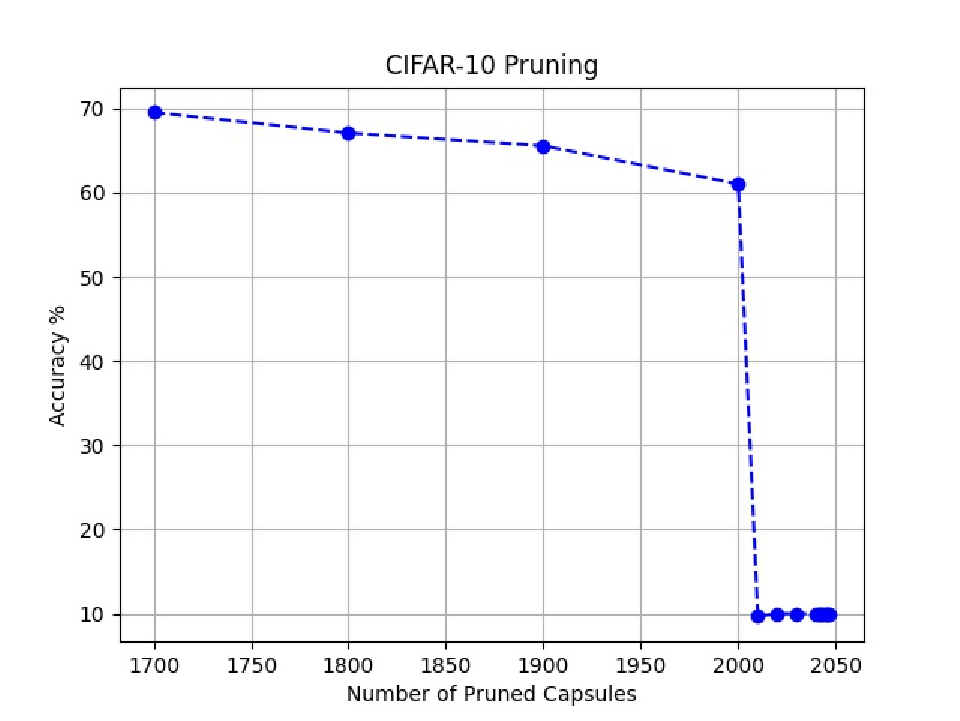}}
\caption{Sudden drop of accuracy on CIFAR-10 dataset compared to MNIST and F-MNIST}
\label{fig:cifar}
\end{figure}

Since CIFAR-10 and SVHN are 32x32 pixels, the starting number of PCs is 2048. We present the performance measures for SVHN in Fig. \ref{fig:svhn}. The starting accuracy for a fully-trained network is 92.65\%. After pruning 1500 PCs out of 2048, we only lose 1.85\% of performance. The results for CIFAR-10 are presented in Fig. \ref{fig:cifar}. The starting and highest possible accuracy for CIFAR-10 is 71.37\% for the original CapsNet architecture. After going through the pruning process, we reach the network's full pruning capacity by removing 2000 PCs. The architecture will not recover if more than 2000 PCs are removed.

\par
\begin{figure}[h!]
\centerline{\includegraphics[scale=0.6]{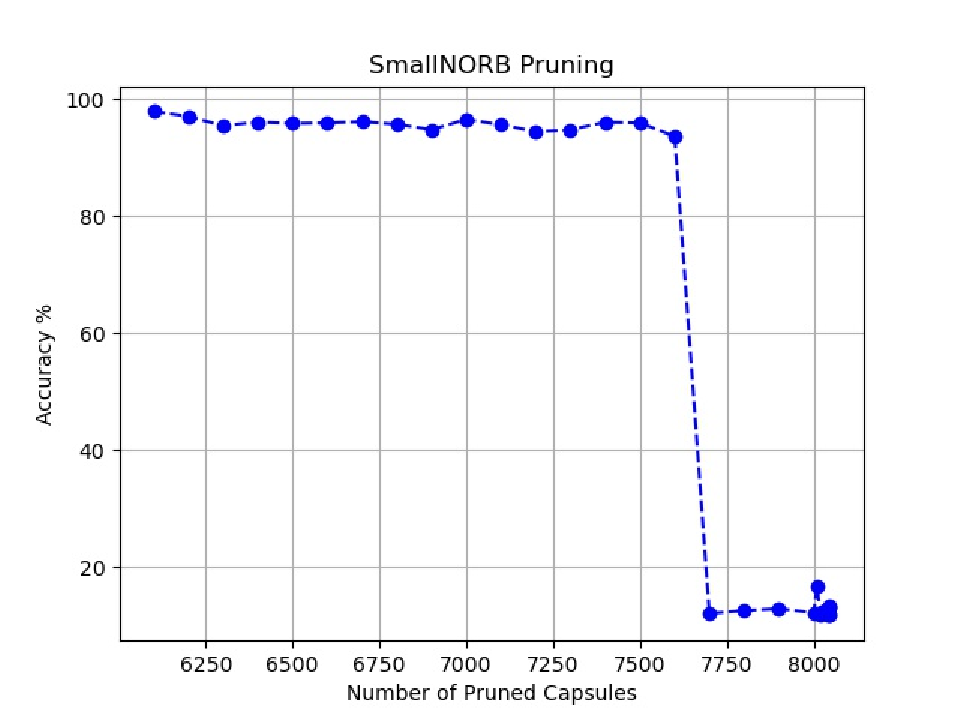}}
\caption{CapsNet cannot recover after pruning 7600 PCs for SmallNORB dataset.}
\label{fig:smallNORB}
\end{figure}

The starting number of PCs for SmallNORB is 8192.The original image size for SmallNORB is 96x96 which results in 51200 number of PCs. We have resized each image to be 48x48 to start with a smaller number of PCs. 
We present the performance measures for SmallNORB in Fig. \ref{fig:smallNORB}. The starting accuracy for this dataset is 94.16\% for the modified 48x48 images. The starting accuracy is the highest achievable training accuracy without overfitting. After pruning 7700 PCs, accuracy fall starts by losing 1.25\%.

\par

\subsection{FLOPS counting}

Floating-point operations are product or summations done by the hardware. The baseline architecture CapsNet performs 276,480 FLOPS to calculate the matrix multiplication, which produces the 1152 PCs for MNIST datasets. This number drops down to 12,480 FLOPS when the number of remaining PCs reaches 52. This 95.48 \% drop in the number of FLOPS is a major improvement over the baseline architecture. Also, Dynamic Routing achieves a 95.36 \% drop in the number FLOPS. 
\par
SVHN and CIFAR-10 datasets experience major drops in their FLOPS count. SVHN experiences a 73.24\% drop in matrix multiplication FLOPS count. This is measured when network is operating on 1500 PCs.
CIFAR-10 can be pruned until it loses 83.01\% of its FLOPS count at the cost of 1.85\% of accuracy. 
SmallNORB also takes advantage of pruning by losing 94.31\% of its FLOPS count. SmallNORB starts at 2.1 M-FLOPS. This can be dropped down to 119,494 FLOPS while removing 7600 PCs.
\par
Since the only part of the network that is changing is the number of PCs (feature extraction and decoder stay untouched), there would be no change in the total number of FLOPS for a single input image in other parts of the architecture.

\subsection{Inference Time Reduction}
\begin{figure}[h!]
\centerline{\includegraphics[scale=0.23]{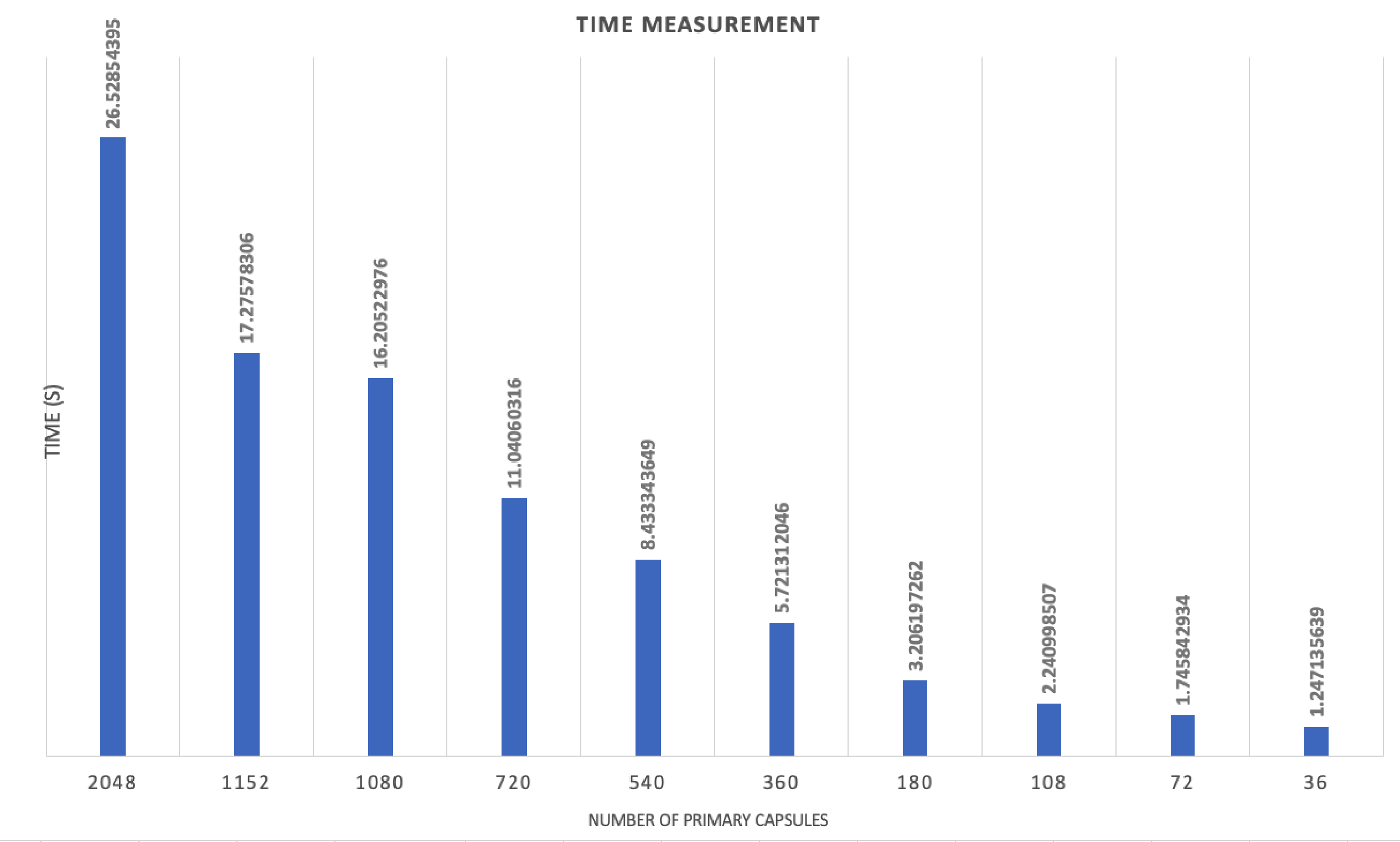}}
\caption{Architecture testing time in seconds for different number of PCs}
\label{fig:time}
\end{figure}

As we showed in the previous section, we save more than 95\% in FLOPS count in the PCs and in the Dynamic Routing section of the architecture. This helps us to lower the inference time. In Fig. \ref{fig:time}, we report the time it takes for CapsNet to output results on 10,000 test images. This is measured when running our experiments on NVIDIA 2080 Ti equipped with 32 GB of memory. 
\par
The pruned architecture produces results 9.90x faster than the original architecture on MNIST categories. The first two bars report for the baseline architecture. The tallest bar reports for 2048 PCs. This is the baseline number of PCs for SVHN and CIFAR-10 datasets. The second bar reports for 1152 PCs which is the baseline number of PCs when the architecture is trained for MNIST category.

\subsection{Discussion}
In this section, we further discuss how pruning affects accuracy, number of FLOPS, and the number of PC in CapsNet and for a number of datasets. In our earlier work \cite{Sharifi2020}, we show that more than 50\% of PC weights are either zero or insignificant. In this work, we report how pruning PCs that are trained on MNIST handwritten digits and Fashion-MNIST does not impact accuracy significantly. The small accuracy drop is due to the fact that most of the PCs are zeros for these two datasets. This is not the case for smallNORB, SVHN and CIFAR-10. These datasets are visually and mathematically \cite{Br2019} more complex compared to MNIST handwritten digits and Fashion MNIST. Also, the images belonging to SVHN and CIFAR-10 are larger in size and 32x32 pixels. In the meantime, images belonging to our modified version of SmallNORB are 48x48 versus the 28x28 images which build MNIST and Fashion MNIST. Our results show that when running on larger and more complex dataset the network needs higher number of remaining PCs to maintain its accuracy. Therefore CapsNet is more sensitive to pruning and cannot recover beyond removing a certain number of PCs for smallNORB, SVHN and CIFAR-10.

\par
This is consistent with our understanding of CapsNet's capacity. Capacity in a deep learning model can be defined by the ability of the network to approximate different functions. A model’s capacity is decided by the volume of trainable parameters it can store. Removing trainable parameters from a network changes its capacity to learn. Since PCs have the highest number of trainable parameters in the CapsNet architecture, they play a significant role in the model’s capacity. As explained earlier smallNORB, SVHN and CIFAR-10 are considered complex datasets compared to MNIST category. Therefore maintaining network accuracy will require a higher number of PCs for SVHN or CIFAR-10 compared to the MNIST category.

\section{Conclusion}
In this paper, we investigated Primary Capsules pruning in CapsNets. CapsNets are a recent generation of image classifiers. Although they have specific advantages over Convolutional Neural Networks, their training and inference phases are inefficient. We trained and fine-tuned the original CapsNet on MNIST handwritten digits, Fashion-MNIST, SVHN, CIFAR-10 and smallNORB datasets. Our results show up to 9.90x speedup and more than 95\% drop in FLOPS over the baseline architecture with minuscule drop of accuracy. We also provided insight into why CapsNet’s behaves differently when pruned on more complex datasets such as SVHN, CIFAR-10 and smallNORB compared to the MNIST and Fashion-MNIST datasets. 

\section*{Acknowledgment}
This research has been funded in part or completely by the Computing Hardware for Emerging Intelligent Sensory Applications (COHESA) project.  COHESA is financed under the National Sciences and Engineering Research Council of Canada (NSERC) Strategic Networks grant number NETGP485577-15.

\bibliographystyle{unsrt}


\end{document}